# Point Cloud Failure Criterion for Composites using k-Nearest Neighbor Classification

Subramaniam Rajan[1], Bilal Khaled[2], and Loukham Shyamsunder[2]
School of Sustainable Engineering & the Built Environment, Arizona State University, AZ 85287

[1]Professor, School of Sustainable Engineering and the Built Environment, Arizona State University, Tempe, AZ

[2]Post-Doctoral Fellow, School of Sustainable Engineering and the Built Environment, Arizona State University, Tempe, AZ

**Abstract**
Numerous theories of failure have been postulated and implemented in various commercial programs for composite materials. Even the best theories have had limited success in predicting damage and failure in validation exercises. In view of this background, many researchers have started exploring the use of multiscale modeling to improve the fidelity of the modeling and simulation of various structural and materials systems. In this paper, a multi-scale modeling scheme is used to illustrate how a combination of virtual and laboratory testing programs can be used to generate a point cloud of failure surface data that can then be queried during finite element analysis at the continuum scale to ascertain if the onset of failure has occurred. The k-nearest neighbor (k-NN) classification concept is used to obtain the answer to the query. A linear, elastic, static finite element example using a unidirectional composite shows that the framework can be generated and used effectively and efficiently with the possibility to extend the approach for all types of composite architectures and behaviors.

## 1.0 Introduction

A typical finite element (FE) analysis (implicit or explicit) of a composite structure at the continuum or structural level involves incremental loading where in every time step (or increment) the state of each finite element in the entire model is updated. The composite architecture is not directly a part of the model and (equivalent) homogenized properties of the composite are used in the analysis. At the FE system level, the calculations can be broken down into at least four components – elastic and inelastic deformations, damage, onset of failure, and post-failure behavior.

Approaches that help define the deformation component can be categorized into several groups [1] - nonlinear elasticity theories, damage theories coupled with elasticity, classical incremental plasticity theory, and endochronic plasticity theory. The damage component captures the degradation of the macroscopic properties. In most cases, damage results in a reduction of load-carrying properties as the effective load transfer mechanisms are altered [2]. Damage can be realized in various ways in composite materials including fiber fracture, matrix cracking, and fiber-matrix debonding [3].

In addition to the complexity of accurately simulating the deformation and damage response of composites, accurately predicting composite failure, particularly in the context of a structural analysis, is an additional challenge [4, 5]. This challenge has been particularly vexing since composites fail in a variety of ways – in compression via fiber crushing, splitting, elastic microbuckling, matrix failure and plastic microbuckling, buckle delamination, and shear band formation [6]; fracture of the material that is brought about by fiber breakage, fiber micro-buckling, fiber pullout, matrix cracking, delamination, debonding or any combination of these mechanisms [7].

There are numerous theories of failure that have been postulated and implemented in various commercial programs. In our earlier paper [4], we examined a few of those that made up part of the suite of theories used in the World-Wide Failure Exercises [8]. However, the organizers of the exercises [9] state that "The designers, wishing to use the models benchmarked in WWFE-II, can only expect a few theories to give acceptable correlation (within ±50%) with test data for 75% of the test cases used." In view of this background, many researchers have started exploring the use of multiscale modeling to improve the fidelity of the modeling and simulation of various structural and materials systems including composite materials [10, 11, 12, 13, 14, 15, 16, 17, 18].

The current research work builds on our prior work with virtual testing [16, 19]. The paper is divided into four sections. First, we discuss the research objectives and assumptions. The next section shows details of how the failure surface is generated. This is followed by details of how the Point Cloud Failure Criterion (PCFC) can be implemented in a continuum level finite element program. Numerical results are then presented and finally, in the concluding remarks section, the research findings are summarized including the shortcomings and how they could possibly be overcome. The T800/F3900 unidirectional composite is used to illustrate the research ideas. Micrographs of the composite are shown in Figure 1.

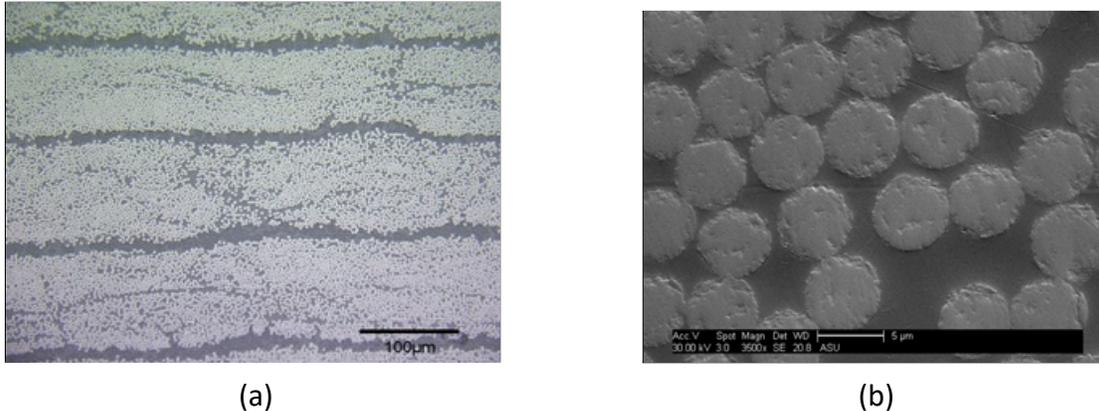

(a)                          (b)

Figure 1. Cross-section of the T800/F3900 composite panel (a) optical micrograph (b) SEM micrograph.

T800/F3900 is a carbon fiber composite with epoxy matrix and a fiber volume fraction of roughly 60%. Experiments show that the composite exhibits brittle failure with little or no post-peak strength [2, 20]. The rich set of available experimentally generated data with this composite and its constituents, makes it an attractive composite to use in the research study.

**2.0 Research Objectives and Assumptions**

The ideas in this paper are illustrated using a simplified example. The components of the overall framework are discussed in detail and how the fundamental ideas can be extended to overcome most of the simplifications are also discussed.

The following are the research objectives and assumptions.

(1) The two-phase composite is assumed to be made of linear, elastic, isotropic materials with brittle failure. Clearly, there are composites that violate this assumption since usually the matrix exhibits elasto-plastic behavior. In our earlier publication dealing with the T800/F3900 composite [21], properties of both the carbon fiber and the fiber/matrix interface were obtained partially through publicly available data and partially through numerical calibration of specific finite element models against experimental T800/F3900 composite data. The FE analyses in our prior studies [2, 4, 5, 29, 20] have been carried out using tension-compression asymmetric, nonlinear orthotropic elasto-plastic rate-dependent behavior with both shell and solid elements. Hence, extending the current model to make it yield the required nonlinear behavior is feasible but computationally expensive.

(2) A linear, elastic, small displacement and small strain analysis is carried out using a planar Representative Volume Element (RVE). A plane strain model in the 2-3 plane is created and analyzed. The stresses and strains in the RVE are averaged to obtain the homogenized values for the entire RVE, for the matrix, and for the fiber.

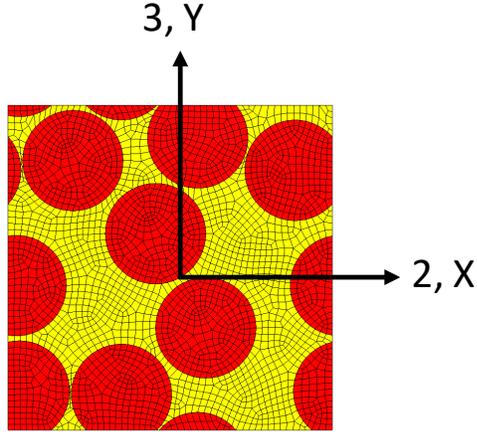

Figure 2. Typical 2-phase RVE showing the fiber (red) and the matrix (yellow). 1-2-3 are the principal material directions.

The finite element model - RVE or Representative Unit Cell (RUC), can be easily extended for capturing a plane stress/strain, or three-dimensional state of stress. The homogenized response of the composite is obtained using a volumetric average of the chosen response variable, e.g. stress and strain, in the fiber and matrix elements in the gage section as [16]

$$\overline{P}_h = \frac{\sum_{j=1}^{e_t} \left( \frac{\sum_{i=1}^{n_{e_t}} \overline{P}_i V_i}{\sum_{i=1}^{n_{e_t}} V_i} \right)_j V_j}{\sum_{j=1}^{e_t} V_j} \qquad (1)$$

where $\overline{P}_h$ is the homogenized material property (stress or strain) at a given time step, $e_t$ is the number of different element types, $n_{e_t}$ is the number of elements in the j$^{th}$ element type, $\overline{P}_i$ is the material property of the i$^{th}$ element. Material property is the average of properties at all integration points, and $V_i$ is the volume of the i$^{th}$ element.

(3) The onset of failure is assumed to occur when the homogenized principal stress in the matrix and/or the fiber exceeds the estimated failure value obtained from experiments of the constituent materials. We define failure onset in a finite element as the state of stress corresponding to a suitably defined peak value, e.g., principal stress, and any further loading may lead to a decrease in the load-carrying capacity of the element (decrease in stress, increase in strain) including sudden, brittle failure. In the numerical examples in this paper, since the behavior is assumed to be linear, the state of stress corresponding to failure onset is obtained simply by scaling the finite element analysis (FEA) stresses to reach the failure value as shown in Eqn. (2).

(4) The RVE is subjected to several combinations of surface tractions, and the finite element analysis results are used to compute the homogenized state of stress for the entire FE model resulting from the failure of either the fiber or the matrix or both. Each generated point $(\sigma_x, \sigma_y, \sigma_z, \tau_{xy}, 0, 0)$ forms a point in the point cloud that represents the failure surface. It should be noted that each point can be transformed, if needed, to a different set of values for different orientations of the x-y coordinate system, e.g., if stresses are needed in the principal material directions.

(5) The initial step in a finite element analysis of a structural system, starts with reading in the stress tensor of all the points in the point cloud failure surface. These points are then stored to be used later with the Approximate Nearest Neighbor algorithm [22, 23, 24]. During the FE analysis, the computed stresses at a Gauss point are used with the ANN algorithm to detect if the current state of stress at the Gauss point is inside or outside the failure envelope. Details are presented in Section 4.0.

**3.0 Generating the Failure Surface**
Generation of the point cloud failure surface is carried out in two steps. In the first step, the RVEs are generated to carry out the finite element analysis. In the second stage, the FEA results are post-processed to compute the point cloud data.

*3.1 Constructing the Representative Volume Elements (RVEs)*
A synthetic two-dimensional (2D) microstructure generation algorithm developed in MATLAB, capable of generating non-overlapping ellipses of varying sizes, aspect ratios, and orientations, was used to generate periodic two-phase microstructures [25, 26]. Fig. 3 shows the basic structure of the RVE.

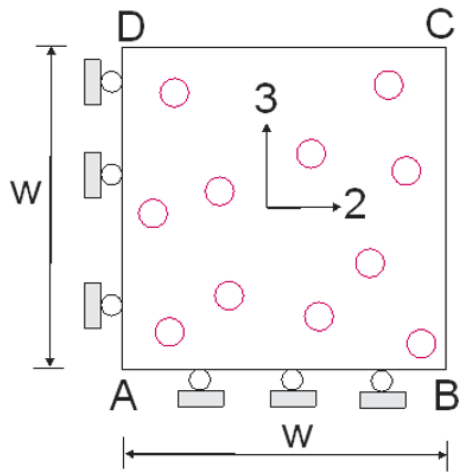

Figure 3. Basic layout of a typical RVE showing the window size, W and the fixity conditions for a plane strain analysis

The circles that represent fiber filaments with a set radius were consecutively added to the window space, W, until the specified area (volume) fraction was obtained. The program created the square representative element area (REA) and inclusion radii in units of pixels, so that the simulations are generalizable and not constrained by physical dimensions. As per the manufacturer[1], each filament (Fig. 1(b)) is approximately 5 µm; hence, 1 pixel translates to ~0.16 µm. To eliminate the possibility of overlapping inclusions, particles were assigned major and minor radii that were 1 pixel larger than the true inclusion size, ensuring that a gap of at least 1 pixel existed between the inclusion edges. Periodic boundary conditions were maintained such that if any part of the inclusion were generated at the edge of the REA, it would wrap around to the other side, including particles located at the corners of the REA. In addition to the images, the microstructural generation algorithm was also used to output a .txt file to export the microstructures into Coreform Trelis™ [27] for meshing prior to FE analysis [Ford et al., 2021].

A convergence analysis was carried out with varying window size W (from 100 to 325 pixels) and varying element size. The auto factor mesh size in Trelis varied from 1 (finest) to 7 (coarsest). An auto factor of 1 produced a mesh where almost all the elements had an aspect ratio of 1. Only the tight regions between fibers produce elements with aspect ratios greater than 1. The FE model consisted of 4-noded quadrilateral finite elements. Edges AD and AB were constrained in the 2 and 3 directions, respectively, and a strain of 0.1% was imposed via displacement boundary conditions on edge BC. The homogenized Young's modulus and Poisson's ratio were compared against the experimentally obtained values of $E_{22}^{exp} = 1.07(10^6) \, psi, v_{23} = 0.439$ [20]. Results of the convergence analysis are shown in Table 1.

Table 1. Convergence Analysis

| Window Size, W | # of models | Results from the Best Model | | | |
|---|---|---|---|---|---|
| | | # of nodes | # of elements | $E_{22}^{FEA}$ (psi) | $\frac{\left(E_{22}^{exp} - E_{22}^{FEA}\right)}{E_{22}^{exp}}$ (%) |
| 100 | 10 | 2949 | 2842 | 1.106(10$^6$) | -3.63 |
| 200 | 10 | 2983 | 2876 | 1.078(10$^6$) | -1.06 |
| 325 | 8 | 3047 | 2937 | 1.200(10$^6$) | -12.48 |

Based on results from Table 1, all subsequent analyses were carried out using a window size of 200.

3.2 Generating the Failure Surface
<u>Finite Element Analysis</u>: Using the values of the optimized parameters (window size and auto factor mesh size), a series of RVEs were generated for FE analysis. For each RVE, varying surface tractions were applied on edges BC and CD as shown in Fig. 4. Results from a sample FE analysis are shown in Fig. 5 (the principal stresses are ordered as $\sigma_1 \leq \sigma_2 \leq \sigma_3$). Each applied surface traction was varied as five discrete values: $(-a, -0.5a, 0, +0.5a, a)$. A total of approximately 120 combinations of surface tractions were generated and used.

---

[1] https://www.toraycma.com/file_viewer.php?id=5126

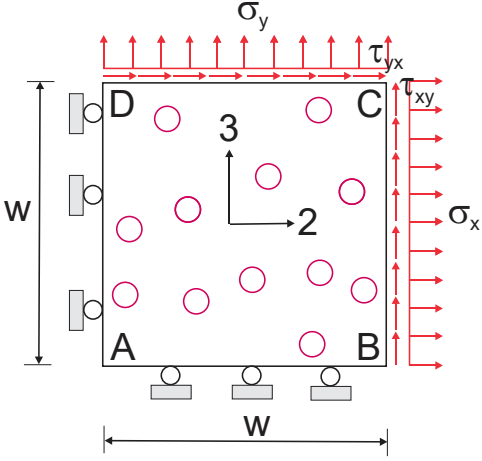
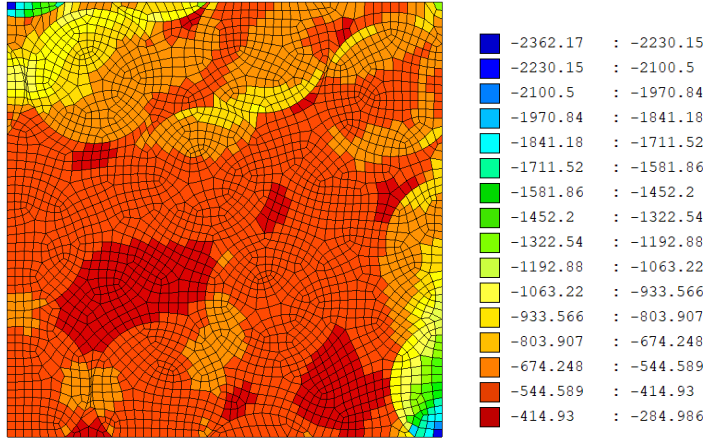

Figure 4. RVE with surface tractions

Figure 5. First principal stress, $\sigma_1$, plot with loading as $(\sigma_x, \sigma_y, \tau_{xy}, \tau_{yx}) = (-a, -a, -a, -a)$

Post-Processing: Once the finite element analysis is completed, the stress tensor from each element is used to compute the three principal stresses from which the maximum tensile stress and the maximum compressive stress are homogenized and computed for (a) the entire RVE, (b) for just the matrix, and (c) for just the fiber, using Eqn. (1). These values are then scaled to find the minimum scaling factor, $s_F$, for the principal stresses to reach the failure surface as

$$s_F = \min_i \left[ (s_F^t)_{fiber}, (s_F^c)_{fiber}, (s_F^t)_{matrix}, (s_F^c)_{matrix} \right], i = 1, 2, \ldots, n_{GP} \tag{2a}$$

where

$$(s_F^t) = \frac{\sigma_F^t}{\sigma_3}, (s_F^c) = \frac{\sigma_F^c}{\sigma_1}, \sigma_3 \neq 0, \sigma_1 \neq 0 \tag{2b}$$

with the superscripts $t, c$ representing tension and compression, $\sigma_F^t, \sigma_F^c$ are the principal failure stresses in tension and compression, and $n_{GP}$ is the total number of Gauss points in the model. The failure stress values are listed in Table 2 along with the material properties for the two materials.

**4.0 Implementing the Failure Criterion**

Unlike an analytical expression for failure, the point cloud data does not readily present the means to ascertain if the state of stress at the Gauss point is inside or outside the failure envelope. In the current work, the Approximate Nearest Neighbor Method [22, 23] is used to estimate if failure has occurred at the stress Gauss point.

The input to the ANN method is a set $P$ of data points in $d$-dimensional space. These points are preprocessed into a data structure, so that given any query point $q$, the nearest (or, generally $k$ nearest) points of $P$ to $q$ can be reported efficiently [23]. Figure 6 shows an example of a teapot represented by a point cloud where the *k*-nearest neighbors are identified once the query point is specified.

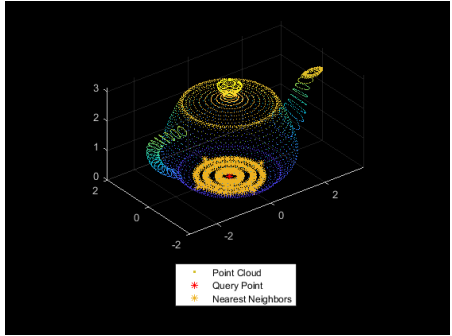 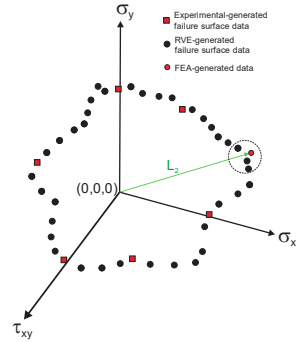

Figure 6. Finding *k*-nearest neighbors in a point cloud[2]

Figure 7. Defining failure surface and *k*-nearest neighbors in the stress-based point cloud

In a similar manner, once the point cloud failure surface is generated either experimentally or via virtual testing (Fig. 7), the nearest neighbors to the FEA-generated stress Gauss point (query point) can be found from which the question can be approximately answered as to whether the FEA-generated point is inside or outside the failure surface (an appropriate value obtained from the nearest neighbors in the surface-generated data, SGD; $\alpha\sigma_{range}$ is a safety factor; $\sigma_{range} = \sigma_{max} - \sigma_{min}, 0 < \alpha < 1$):

$$\text{Outside}: L_2^{FEA} \geq L_2^{SGD} - \alpha\sigma_{range}$$
$$\text{Inside}: L_2^{FEA} < L_2^{SGD} - \alpha\sigma_{range} \quad (3a)$$

where for vector **v**

$$L_2 = \sqrt{\sum_{i=1}^{d} v_i^2} \quad (3b)$$

For stress-based failure surface data, $d = 3, 4, 6$ for plane stress, plane strain and full three-dimensional state of stress, respectively. In fact, a very general Minkowski $L_r$ norm can be used

$$L_r = \left(\sum_{i=1}^{d} |v_i|^r\right)^{1/r} \quad (3c)$$

While the appropriate value obtained from the nearest neighbors can take on many forms, in this paper, the average value, $L_2^{avg} = \dfrac{\sum_{i=1}^{k}\left(L_2^{SGD}\right)_i}{k}$ is used.

Use of the ANN library in a C++ computer program is relatively simple. First, *P*, *d* and the error bound, $\varepsilon$ are specified. Next, the data points are passed to the library after which the ANN data structure is created. These steps are done only once after which the query point, $q$, can be specified as many times as required with *k* (number of nearest points) as an additional input. The efficiency and accuracy of the ANN method is a function of (a) the number of points used to create the failure surface, P, (b) how well the points are distributed in the search space so as yield sufficiently close points during the search process, (c) the accuracy, $\varepsilon$, with which the nearest neighbors need to be located, and (d) the number of nearest neighbors needed to accurately approximate the failure surface. The accuracy [23] is defined in terms of an optional real value $\varepsilon \geq 0$. If supplied, then the $i^{th}$ nearest neighbor is a $(1+\varepsilon)$ approximation to the true $i^{th}$ nearest neighbor. That is, the true (not squared) distance to this point may exceed the true distance to the real $i^{th}$ nearest neighbor of $q$ by a factor of $(1+\varepsilon)$. If $\varepsilon$ is omitted, then nearest neighbors are computed exactly.

---

[2] https://www.mathworks.com/help/vision/ref/pointcloud.findnearestneighbors.html

## 5.0 Numerical Results

The proposed framework is created and used to illustrate and validate the ideas. The material properties of the matrix and fiber are shown in Table 2. The overall framework and steps are shown in Fig. 8.

Table 2. Material Properties

| Material | Young's Modulus (psi) | Poisson's Ratio | Principal Failure Stress, Tension (psi) | Principal Failure Stress, Compression (psi) |
|---|---|---|---|---|
| T800 (fiber) | 2.25(10$^6$) | 0.25 | 35000 | 35000 |
| F3900 (matrix) | 4.09(10$^5$) | 0.387 | 15375 | 23000 |

A specialized MATLAB program based on a synthetic microstructure generator [26] is used to create the RVEs, i.e. nodes and elements. The AUTOML program is used to read the FE model and add boundary conditions – fixity conditions and applied surface tractions, thus creating all the FE models required to generate the point cloud failure surface. The surface tractions are automatically generated by varying them in equally spaced intervals in the range $-1000 \leq \sigma_{ij} \leq 1000$. The ANN failure surface database (FSD) is created for use by another program, FAILSURF where verification and validation take place.

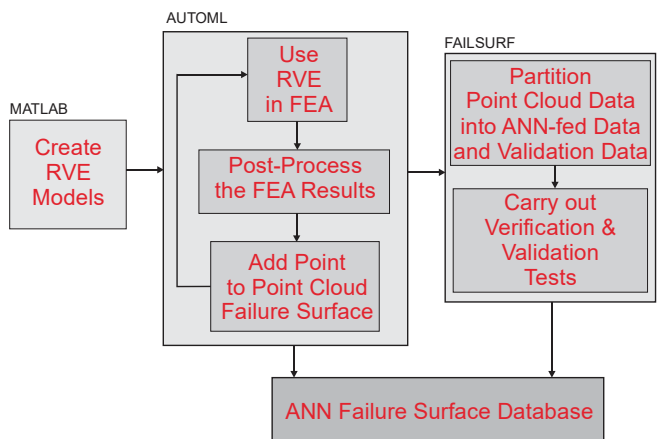

Figure 8. Overall software framework

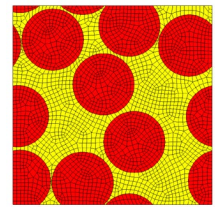

(a) E22NU23_vf60-139

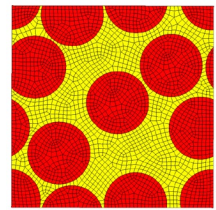

(b) E22NU23_vf60-176

Figure 9. RVEs used in this study

**First Validation Test**: In this study, only two arbitrarily selected microstructure from the hundreds of generated microstructures from our earlier study [28] are used as shown in Fig. 9(a)-(b). A total of *2n* data points are generated on the failure surface.

In the first validation test, the *2n* data points are broken into two parts – 80% of the data points selected randomly, are stored in the ANN database, and the rest of the data points are used in the validation tests by the FAILSURF program. Table 3 shows a select few sample points for *n=124*.

Table 3. RVE-generated Points on the Failure Surface (Select few shown)

| Model (n=124) | Failure Surface Stress State $\{\sigma_x, \sigma_y, \sigma_z, \tau_{xy}\}$ psi | Failure Due to |
|---|---|---|
| 139: Run1 | $\{-22155, -21989, -13190, -4765\}$ | Matrix compression |
| 139: Run100 | $\{9910, -28502, -5485, -6584\}$ | Fiber compression |
| 139: Run59 | $\{19946, -19967, 33, 12909\}$ | Matrix tension |
| 139: Run80 | $\{362, -27351, -8008, -9215\}$ | Matrix compression |
| 139: Run99 | $\{11041, 16510, 8222, 3636\}$ | Matrix tension |
| 176: Run101 | $\{17599, -29816, -3676, -3874\}$ | Fiber compression |
| 176: Run14 | $\{-28359, 9355, -5713, 6195\}$ | Matrix compression |
| 176: Run35 | $\{605, -1392, -617, 26658\}$ | Fiber tension |
| 176: Run90 | $\{-445, 1024, 454, -19604\}$ | Matrix tension |
| 176: Run61 | $\{-18448, -17894, -10705, -11713\}$ | Fiber compression |

The results from three data space sizes are shown in Table 4. The terms used in the table are explained below.

*Distribution*: To gage how well the generated points are distributed in the ANN search space, the search space is divided into bins (similar to quadrants in 2D space and octants in 3D space). There are 16 bins $(2^4)$ for the plane strain models, and the number of data points in each bin is determined after the FEA. Min, max and avg refer to the minimum, maximum and average number of points considering all the bins, and zero is the number of bins with no data points.

*Accuracy*: The accuracy of the prediction framework can be found by carrying out the validation tests. A binary evaluation is carried out – accurate if the prediction matches the answer (data point is either inside or outside the failure envelope), an error otherwise.

*Wall Clock Time*: This is the total time taken to execute the AUTOML and FAILSURF programs. The runs were made on a Dell Precision T5610 workstation running 64-bit Windows 10 Enterprise with Intel Xeon E5-2637 CPU and 24 GB RAM. The programs were run sequentially using only one CPU core at a time.

Table 4. Summary of the Test Runs

| Data Space Size, 2n | Bin Distribution (min, max, avg, zero) | ANN Parameters $(\alpha, \varepsilon, k)$ | # of Validation Tests | Accuracy (%) | Wall Clock Time (s) |
|---|---|---|---|---|---|
| 248 | (7, 46, 20.7, 4) | (0.001, 0, 4) | 50 | 43 | 499 |
| 248 | (7, 46, 20.7, 4) | (0.01, 0, 4) | 50 | 68 | 499 |
| 248 | (7, 46, 20.7, 4) | (0.1, 0, 4) | 50 | 100 | 499 |
| 248 | (7, 46, 20.7, 4) | (0.001, 0, 3) | 50 | 48 | 499 |
| 248 | (7, 46, 20.7, 4) | (0.01, 0, 3) | 50 | 70 | 499 |
| 248 | (7, 46, 20.7, 4) | (0.1, 0, 3) | 50 | 100 | 499 |
| 684 | (21, 130, 57, 4) | (0.001, 0, 4) | 136 | 54 | 1254 |
| 684 | (21, 130, 57, 4) | (0.01, 0, 4) | 136 | 76 | 1254 |
| 684 | (21, 130, 57, 4) | (0.1, 0, 4) | 136 | 100 | 1254 |
| 684 | (21, 130, 57, 4) | (0.001, 0, 3) | 136 | 51 | 1254 |
| 684 | (21, 130, 57, 4) | (0.01, 0, 3) | 136 | 77 | 1254 |
| 684 | (21, 130, 57, 4) | (0.1, 0, 3) | 136 | 100 | 1254 |
| 1456 | (47, 279, 121.3, 4) | (0.001, 0, 4) | 292 | 50 | 2594 |
| 1456 | (47, 279, 121.3, 4) | (0.01, 0, 4) | 292 | 76 | 2594 |
| 1456 | (47, 279, 121.3, 4) | (0.1, 0, 4) | 292 | 100 | 2594 |
| 1456 | (47, 279, 121.3, 4) | (0.001, 0, 3) | 292 | 55 | 2594 |
| 1456 | (47, 279, 121.3, 4) | (0.01, 0, 3) | 292 | 75 | 2594 |
| 1456 | (47, 279, 121.3, 4) | (0.1, 0, 3) | 292 | 100 | 2594 |
| 2660 | (127, 618, 249.4, 0) | (0.001, 0, 4) | 533 | 65 | 5022 |
| 2660 | (127, 618, 249.4, 0) | (0.01, 0, 4) | 533 | 97 | 5022 |
| 2660 | (127, 618, 249.4, 0) | (0.1, 0, 4) | 533 | 100 | 5022 |
| 2660 | (127, 618, 249.4, 0) | (0.001, 0, 3) | 533 | 68 | 5022 |
| 2660 | (127, 618, 249.4, 0) | (0.01, 0, 3) | 533 | 95 | 5022 |
| 2660 | (127, 618, 249.4, 0) | (0.1, 0, 3) | 533 | 100 | 5022 |

The relationships between the different parameters in Table 4 are shown in Fig. 10.

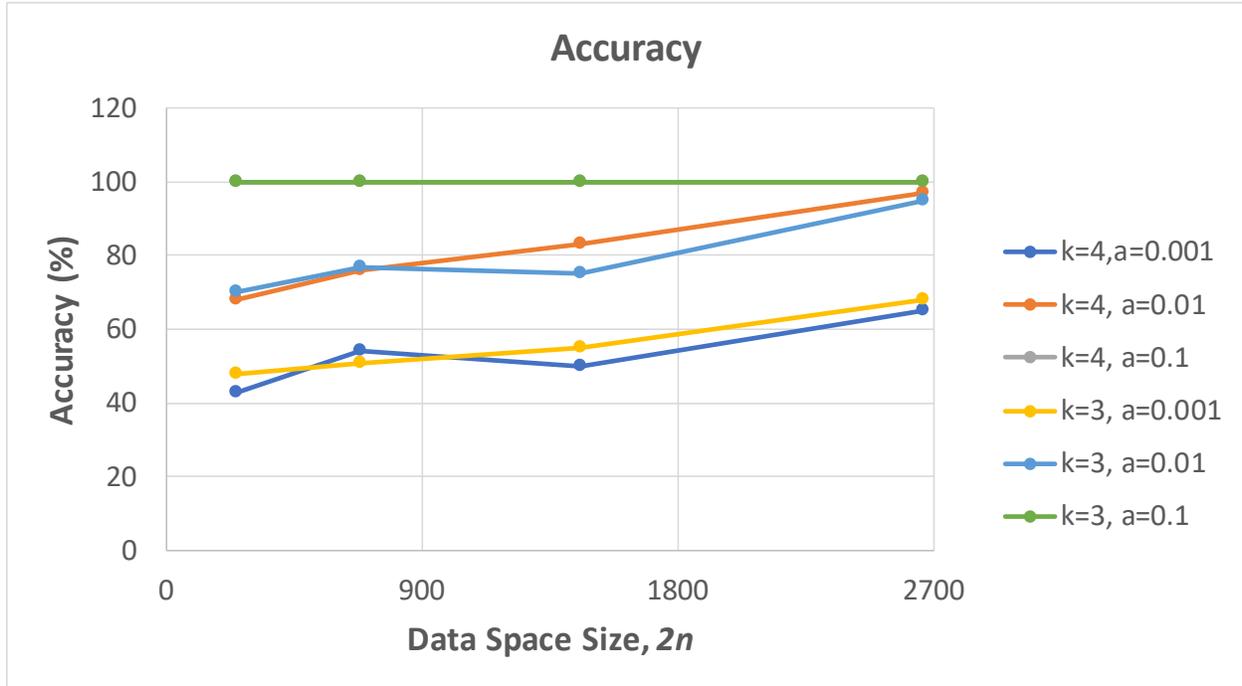

Figure 10. Accuracy as a function of P=*1.6n*, k, $\alpha$

*Discussions*: The results indicate that acceptable predictions can be obtained using the developed framework.

(1) The results indicate that the accuracy is a function of several parameters – the size of the data space, *n*, the safety factor, $\alpha$, and the number of neighboring points, *k*. The accuracy parameter, $\varepsilon$, was not varied as the decrease in computing time with increasing $\varepsilon$ (to compute approximate nearest neighbor) was imperceptible.

(2) The binary evaluation in accuracy is perhaps too harsh. However, as one can see, a minimal increase in $\alpha$ can greatly increase the prediction accuracy.

(3) With the way the data points are generated, some bins may have no points. For example, it is not possible to have $\{+\sigma_x, +\sigma_y, -\sigma_z, +\tau_{xy}\}$ state of stress on the failure surface until *n=1330*.

(4) The nearest neighbor can take multiple forms – *k* nearest neighbors, neighbors within a specified radius, computing weighted average, $L_2^{SGD}$ from the *k* nearest neighbors, etc. However, using a simple average appears to work well with the examples considered in this paper.

**Second Validation Test**: In this second validation test, the ANN database created from *2n=2660* models from the first validation test is used. However, the validation tests involve a different RVE (same fiber and matrix properties, same volume fraction, but different fiber arrangement) that is shown in Fig. 11. This validation test is divided into two parts.

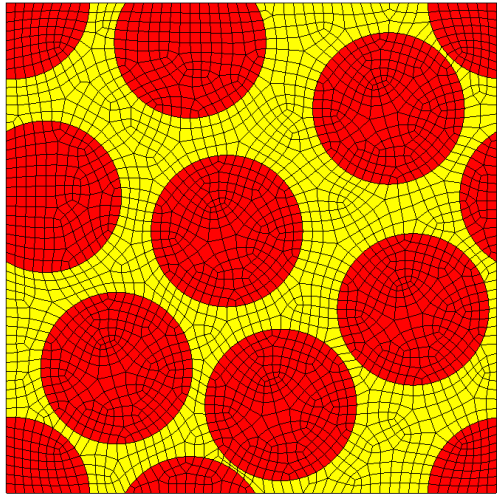

Figure. 11. E22NU23_vf60-160

**Part A**: The points on the failure surface from this model (E22NU23_vf60-160) are queried using the ANN database to determine if the point is inside or outside the failure surface. The summary of this exercise is shown in Table 5. An additional value, $\alpha = 0.05$, is used, and the Wall Clock Time represents only the time spent in the FAILSURF program.

Table 5. Summary of the Validation Runs (Part A)

| Data Space Size, $2n$ | ANN Parameters $(\alpha, \varepsilon, k)$ | # of Validation Tests | Accuracy (%) | Wall Clock Time (s) |
|---|---|---|---|---|
| 2660 | (0.001, 0, 4) | 124 | 18 | 1 |
| 2660 | (0.01, 0, 4) | 124 | 53 | 1 |
| 2660 | (0.05, 0, 4) | 124 | 100 | 1 |
| 2660 | (0.1, 0, 4) | 124 | 100 | 1 |
| 2660 | (0.001, 0, 3) | 124 | 18 | 1 |
| 2660 | (0.01, 0, 3) | 124 | 50 | 1 |
| 2660 | (0.05, 0, 3) | 124 | 100 | 1 |
| 2660 | (0.1, 0, 3) | 124 | 100 | 1 |
| 2660 | (0.001, 0, 4) | 1330 | 18 | 1 |
| 2660 | (0.01, 0, 4) | 1330 | 49 | 1 |
| 2660 | (0.05, 0, 4) | 1330 | 99 | 1 |
| 2660 | (0.1, 0, 4) | 1330 | 100 | 1 |
| 2660 | (0.001, 0, 3) | 1330 | 20 | 1 |
| 2660 | (0.01, 0, 3) | 1330 | 53 | 1 |
| 2660 | (0.05, 0, 3) | 1330 | 99.6 | 1 |
| 2660 | (0.1, 0, 3) | 1330 | 100 | 1 |

**Part B**: In the second part, the failure data from the E22NU23_vf60-160 RVE are divided into two parts – one-third of the failure test data points are left intact while the stress tensor in the remaining two-thirds is randomly scaled by a factor, $f$, with $0.5 \leq f \leq 1.2$. If $f < 1$, the point is shifted to inside the failure surface, otherwise it is shifted outside. The primary objective of this exercise is to see how the developed algorithm would work in an actual nonlinear analysis where the FE analysis is carried out incrementally with the stresses increasing to approach the failure surface from inside. A few new terms are defined first.

*Correct Prediction*: A prediction is counted as being correct if the prediction matches the data, i.e. failure prediction for test data points on or outside the failure surface, and no failure prediction otherwise.

*False Positive*: A false positive is the test case where the test data point is classified as being outside the failure envelope, but the test data point is in fact inside the failure envelope.

*False Negative:* A false negative is the test case where the test data point is classified as being inside the failure envelope, but the test data point is in fact outside the failure envelope.

To get a sense of how far these false positive and negative predictions are from the ANN SGD failure surface, the min. and max. distance prediction errors are computed where the error is defined as

$$error(\%) = \frac{L_2^{test} - L_2^{avg}}{\sigma_{range}} \times 100 \tag{4}$$

Table 6. Summary of the Validation Runs (Part B). (2n=2660)

| ANN Parameters $(\alpha, \varepsilon, k)$ | # of Correct Predictions (% correct) | # of False Positives (FP) | [Min, Max] Distance Prediction FP Error (%) | # of False Negatives (FN) | [Min, Max] Distance Prediction FN Error (%) |
|---|---|---|---|---|---|
| (0.001, 0, 4) | 752 (57) | 4 | [0, 1.7] | 574 | [-5.7, 0.1] |
| (0.01, 0, 4) | 965 (73) | 11 | [-0.9, 0.4] | 354 | [-5.7, -1] |
| (0.05, 0, 4) | 1204 (91) | 119 | [-4.9, -0.1] | 7 | [-5.7, -5] |
| (0.1, 0, 4) | 1079 (81) | 251 | [-9.9, 0.7] | 0 | [NA, NA] |
| (0.001, 0, 3) | 760 (57) | 4 | [0, 1.6] | 566 | [-5.7, -0.1] |
| (0.01, 0, 3) | 992 (75) | 11 | [-1, 1.3] | 327 | [-5.7, -1] |
| (0.05, 0, 3) | 1211 (91) | 115 | [-5, 1.5] | 4 | [-5.7, -5] |
| (0.1, 0, 3) | 1059 (80) | 271 | [-10, 1.1] | 0 | [NA, NA] |

*Discussions*:
(1) Part A: The results indicate that an acceptable prediction can be obtained using the developed framework. The time required to carry out the prediction using the ANN database is insignificant. From the obtained results, it appears that a 5% safety factor is adequate for an accurate prediction.
(2) Part B: While neither false positive nor false negative predictions are desirable, from a conservative design perspective, a false positive is preferable to a false negative. Increasing the safety factor decreases the number of false negatives. However, making it large increases the number of false positives while decreasing the number of false negatives. It appears as if, even with this example, a 5% safety factor yields the best results.

**6.0 Conclusions**

A new failure criterion for composites is proposed. Given the fact that composites are complex both in terms of the underlying constituent materials as well as the modes of failure, an experimentally/numerically generated failure envelope provides a convenient way of overcoming the challenges of generating analytical expressions for various failure mechanisms and generating the failure data in the laboratory. While the computational cost of constructing the failure surface is a function of the complexity of the RVE model, the cost of finding if a state of stress is inside or outside the failure envelope is by contrast miniscule and is of the same order of magnitude as using analytical expressions. Clearly the RVE can be subjected to multi-axial states of stress that is extremely difficult to carry out in a laboratory.

A few points to note as further investigations take place to improve the applicability of the proposed failure criterion.

(1) The framework can be used for carrying out either a nonlinear implicit or explicit FEA with obviously increasing computational cost. However, it should be noted that the FEA cost is manageable especially given the ready access and availability of CPU cores. Moreover, it should be noted that the hundreds and thousands of models that need to be executed can be executed individually (in parallel) – an example of embarrassingly parallel problem, and this task needs to be done once and only once for a given material.

(2) Almost 99% of the computational time is taken in generating the failure surface in the AUTOML program and very little time is taken in the generating the ANN database and launching the hundreds of queries. Given the fact that in actual structural FEA there are possibly hundreds and thousands of Gauss points and a very large number of time steps, the ANN-based prediction compute cost is likely to be marginally more expensive than using analytical expressions while providing a highly enhanced accuracy. Our preliminary runs show that about a million queries can be carried out in about a second on the computing platform used in this study.

(3) The framework can be easily adapted to carry out displacement-controlled tests including handling strain softening behavior seen with some composites. A strain-based failure surface can be constructed to predict when the element should be eroded from the FE model during an explicit FEA.

**CRediT authorship contribution statement**
**Subramaniam Rajan**: Methodology, Validation, Software, Investigation, Formal analysis, Writing - original draft, Data curation, Visualization. Project administration, resources, funding acquisition.
**Bilal Khaled**: Experimental analysis, Validation, Formal analysis, Writing, Visualization.

**Loukham Shyamsunder**: Validation, Formal analysis, Writing, Visualization.

**Acknowledgements**

The authors gratefully acknowledge the support of the Federal Aviation Administration through Grant #12-G-001 titled "Composite Material Model for Impact Analysis" and #17-G-005 titled "Enhancing the Capabilities of MAT213 for Impact Analysis", William Emmerling and Dan Cordasco, Technical Monitors.

**Declaration of Competing Interest**

The authors declare that they have no known competing financial interests or personal relationships that could have appeared to influence the work reported in this paper.

**Data Availability**

The raw/processed data required to reproduce these findings is available from the corresponding author.